\documentclass[runningheads]{llncs}
\usepackage[T1]{fontenc}
\usepackage{booktabs}
\usepackage{graphicx}
\usepackage[misc]{ifsym}
\newcommand{\corr}{(\Letter)}
\usepackage[nolist]{acronym}
\usepackage[english]{babel}
\usepackage{randtext}
\usepackage{algorithm}
\usepackage{algpseudocode}
\usepackage{xcolor}
\usepackage{rotating}
\usepackage{cite}

\usepackage{multirow}
\usepackage{multicol}
\usepackage{booktabs}
\usepackage{threeparttable}
\usepackage{graphicx}

\usepackage{nicefrac}

\usepackage{amsmath}
\usepackage[mode=buildnew]{standalone}
\usepackage{pgfplots}
\pgfplotsset{compat=1.18}
\begin{document}
\title{YANA: Bridging the Neuromorphic Simulation-to-Hardware Gap}
\author{
    Brian Pachideh\inst{1,3} \corr \and
    Sven Nitzsche\inst{1,3} \and
    Moritz Neher\inst{1,2,3} \and
    Jann Krausse\inst{2,3} \and
    Carmen Weigelt\inst{2,3} \and
    Klaus Knobloch\inst{2} \and
    Victor Pazmino Betancourt\inst{1,3} \and
    Juergen Becker\inst{1,3}
}
\authorrunning{Brian Pachideh et al.}
\institute{
    FZI Research Center for Information Technology, Karlsruhe, Germany \\ \email{\{pachideh, nitzsche, neher, pazmino\}@fzi.de}
    \and
    Infineon Technologies, Dresden, Germany \\
    \email{\{jann.krausse, carmen.weigelt, klaus.knobloch\}@infineon.com}
    \and
    Karlsruhe Institute of Technology, Karlsruhe, Germany \\ \email{becker@kit.edu}
}
\maketitle
\begin{acronym}[Longest Abrev]
\acro{ANN}{Artificial Neuronal Network}
\acro{AE}{Auto-Encoder}
\acro{CNN}{Convolutional Neural Network}
\acro{DNN}{Deep Neural Network}
\acro{LSTM}{Long Short-Term Memory}
\acro{RNN}{Reccurent Neural Network}
\acro{NN}{Neural Network}
\acro{ML}{Machine Learning}
\acro{DL}{Deep Learning}
\acro{AI}{artificial intelligence}
\acro{DFT}{Discrete Fourier Transform}
\acro{FFT}{Fast Fourier Transform}
\acro{STFT}{Short-Time Fourier transform}
\acro{BPTT}{Backpropagation Through Time}
\acro{BP}{Backpropagation}
\acro{LMD}{Local Mean Decomposition}
\acro{GRF}{Gaussian Receptive Field}
\acro{TTFS}{Time-to-First-Spike}
\acro{PCA}{Principal component analysis}

\acro{FPGA}{field-programmable gate array}
\acro{US+}{UltraScale+}
\acro{SoC}{system on chip}
\acro{NoC}{network on chip}
\acro{SoM}{system on module}
\acro{LSb}{least significant bit}
\acro{PL}{programmable logic}
\acro{PS}{processing system}
\acro{RTL}{register transfer level}
\acro{CU}{control unit}
\acro{LUT}{lookup table}
\acro{NC}{Neuromorphic Computing}
\acro{SNN}{Spiking Neural Network}
\acro{LSNN}{Long Short-Term Spiking Neural Network}
\acro{LIF}[LIF]{Leaky Integrate-and-Fire}
\acro{IF}[IF]{Integrate-and-Fire}
\acro{LI}[LI]{Leaky Integrate}
\acro{ALIF}{Adaptive Leaky Integrate-and-Fire}
\acro{adex}[AdEx]{adaptive exponential integrate-and-fire}

\end{acronym}

\begin{abstract}
    Spiking Neural Networks (SNNs) promise significant advantages over conventional Artificial Neural Networks (ANNs) for applications requiring real-time processing of temporally sparse data streams under strict power constraints -- a concept known as the Neuromorphic Advantage. However, the limited availability of neuromorphic hardware creates a substantial simulation-to-hardware gap that impedes algorithmic innovation, hardware-software co-design, and the development of mature open-source ecosystems. To address this challenge, we introduce Yet Another Neuromorphic Accelerator (YANA), an FPGA-based digital SNN accelerator designed to bridge this gap by providing an accessible hardware and software framework for neuromorphic computing.
YANA implements a five-stage, event-driven processing pipeline that fully exploits temporal and spatial sparsity while supporting arbitrary SNN topologies through point-to-point neuron connections. The architecture features an input preprocessing scheme that maintains steady event processing at one event per cycle without buffer overflow risks, and implements hardware-efficient event-driven neuron updates using lookup tables for leak calculations. We demonstrate YANA's sparsity exploitation capabilities through experiments on the Spiking Heidelberg Digits dataset, showing near-linear scaling of inference time with both spatial and temporal sparsity levels. Deployed on the accessible AMD Kria KR260 platform, a single YANA core utilizes 740 LUTs, 918 registers, 7 BRAMS and 24 URAMs, supporting up to $2^{17}$ synapses and $2^{10}$ neurons. We release the YANA framework as an open-source project, providing an end-to-end solution for training, optimizing, and deploying SNNs that integrates with existing neuromorphic computing tools through the Neuromorphic Intermediate Representation (NIR).
\end{abstract}
\section{Introduction}
    \subsubsection{On the Promises of the Neuromorphic Advantage.}
\ac{NC} and \acp{SNN} hold the promise of improving over conventional \acp{ANN} in certain AI applications that require real-time processing of temporally sparse data streams, especially under strict power constraints~\cite{bos_sub-mw_2022, shrestha_efficient_2024}. This concept is broadly referred to as the \textit{Neuromorphic Advantage}~\cite{diamond_comparing_2016,aimone_provable_2021,kudithipudi_neuromorphic_2025}. The expected advantage is rooted in algorithmic properties inherent to \acp{SNN}, such as the sparsity in neuronal activations and connectivity, spatially and temporally sparse spike input encoding or increased expressivity through brain-inspired features like synaptic delays or some forms of online adaptivity.

\subsubsection{The Neuromorphic Simulation-to-Hardware Gap.}
To realize the advantage, SNNs need to be deployed on neuromorphic hardware platforms which exploit the aforementioned algorithmic properties via its hardware architecture and SNN-specific processing elements.
Architectural features that support SNNs include event-driven computation, massive parallelism, co-location of memory and processing (reducing the von Neumann bottleneck), inherent scalability, and novel mixed signal or asynchronous digital circuit designs~\cite{richter_dynap-se2_2024,hoppner_spinnaker_2022,davies_taking_2021}.
Despite the compelling promise of \ac{NC}, the practical and widespread realization of the Neuromorphic Advantage faces hurdles due to the limited availability and accessibility of event-driven neuromorphic accelerators. This scarcity creates a \textit{simulation-to-hardware gap}: while many innovative \ac{SNN} algorithms and \ac{NC} concepts demonstrate considerable promise in simulated environments, their empirical validation, performance characterization, and iterative refinement on actual neuromorphic hardware are severely constrained. This gap, in turn, impedes:
\begin{itemize}
    \item \textbf{HW/SW Co-Design Cycle:} The feedback loop, where insights from hardware performance inform algorithmic refinement and new hardware designs \cite{kudithipudi_neuromorphic_2025}.
    \item \textbf{Algorithmic Innovation:} Without access to real hardware, the exploration of novel SNN architectures, training methods, and optimization techniques is often limited to theoretical constructs or simulations that may not translate well to physical devices.
    \item \textbf{Open-Source Ecosystem Maturation:} A robust open-source software ecosystem lowers the entry barrier to neuromorphic research and development. Without broad access to neuromorphic hardware, the neuromorphic equivalents to programming models, compilers, and interoperable APIs cannot mature effectively~\cite{muir_road_2025}.
    \item \textbf{Hardware-Grounded Benchmarking:} Establishing clear, fair, and widely accepted benchmarks that assess performance on actual neuromorphic hardware (rather than just simulation) becomes exceedingly difficult, making objective comparisons between approaches challenging~\cite{yik_neurobench_2025}.
\end{itemize}
\subsubsection{Related Work.}
Over several decades of neuromorphic hardware acceleration research, the field has recently reached integration levels of billions of neurons and synapses on a single chip. The recent review by Kudithipudi et al.~\cite{kudithipudi_neuromorphic_2025} provides a comprehensive overview of the different architectures of neuromorphic accelerators. However, almost all of these systems remain unavailable to the public, serving to close the simulation-to-hardware gap for only a small number of researchers and developers. While a few systems are commercially available, they are heavily constrained in their programmability to meet strict power and area requirements for commercial viability. To properly close the simulation-to-hardware gap and enable rapid innovation in the public research space, we believe it is necessary to provide an open-source framework that allows the development of neuromorphic hardware accelerators that are both programmable and accessible to a broad community.
Recent related work addresses the acceleration of \ac{DL} programmed \acp{SNN} onto \acp{FPGA}, backed by open-source code repositories including \ac{RTL} descriptions and integration with one \ac{DL} training library out of the neuromorphic open-source ecosystem~\cite{leone_syntzulu_2025,carpegna_spiker_2024,saulquin_modnef_2025}.
At first glance, these works appear to provide a solution to the simulation-to-hardware gap, as they enable the deployment of \ac{SNN} models onto \ac{FPGA} platforms. However, these existing hardware architectures primarily support \ac{SNN} graphs that are constrained by the layered, sequential connectivity patterns inherent to conventional \ac{DL} programming models, making them inadequately prepared for arbitrary, highly recurrent \ac{SNN} graphs that may emerge from future neuromorphic programming paradigms. Furthermore, the aforementioned accelerators implement clock-driven updating mechanisms rather than event-driven processing, limiting their ability to exploit both spatial and temporal sparsity inherent in \ac{SNN} computations.
Overall, there is a lack of open-source, event-driven neuromorphic accelerators that are designed to support arbitrary \ac{SNN} graphs and that can be used to close the simulation-to-hardware gap. This gap hinders the development of neuromorphic hardware accelerators that can fully exploit the Neuromorphic Advantage.

\subsubsection{Our Contribution.}
To address the availability of event-driven, open-source frameworks that provide \ac{SNN} acceleration and thus bridge the simulation-to-hardware gap, we introduce Yet Another Neuromorphic Accelerator (YANA). YANA is an FPGA-based digital SNN accelerator that processes and communicates event-by-event, fully exploiting sparsity in activations, SNN connectivity, and event packet transfer. 
YANA leverages the near-memory computing paradigm and its architecture supports the acceleration of arbitrary SNN graphs, providing flexibility beyond the structures commonly found in the \ac{DL} programming model.

We present hardware resource utilization metrics for a single YANA core and this work's experimental prototype. We also demonstrate how YANA's event-driven architecture can exploit temporal and spatial sparsity to reduce inference latency by deploying a suite of \acp{SNN} with varying pruning levels and input sparsities on the Spiking Heidelberg Digits (SHD) \cite{cramer_heidelberg_2022} dataset.

Furthermore, we release the first public open-source version of the YANA framework for the training, optimization, and deployment of SNNs onto a single YANA core.
\footnote[1]{https://github.com/pachideh/yana-prototype}

\section{YANA Hardware Accelerator}
    \label{sec:yana_hw_accelerator}
    \subsection{YANA Core Architecture}
YANA is a digital \ac{SNN} emulator, as all of its logic and memory resources are dedicated either to implementing neurosynaptic operations or to transporting events. Furthermore, YANA is an event-driven processor; it only processes events once they arrive and is otherwise idle. It implements a near-memory processing pipeline that encapsulates multiple neurons and synapses. The neurons and synapses share the same processing logic within the core in time-multiplexed fashion, which reduces resource utilization~\cite{frenkel_bottom-up_2023}. Furthermore, processing and communication resources operate concurrently -- neuron state updates, spike emission and spike transport all occur in parallel within a timestep.
\begin{figure*}
  \centering
  \makebox[\textwidth]{\includegraphics[width=0.95\textwidth]{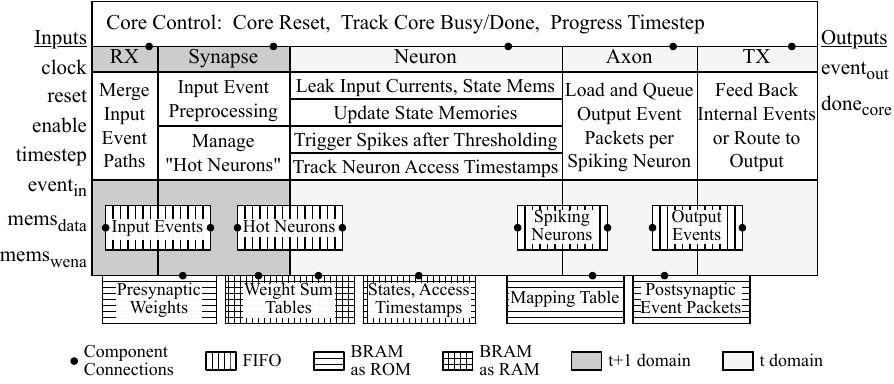}}
  \caption{High-level block diagram of the YANA core.}
  \label{YANA-core-block-diagram}
\end{figure*}
The core's processing pipeline consists of five main stages as depicted in Figure~\ref{YANA-core-block-diagram}. The stages are briefly described as follows:
\begin{itemize}
  \item \textbf{Input Stage (RX):}
  The input stage receives events from two possible stream sources, either from the external interconnect or from the core's internal feedback path.
  \item \textbf{Synapse Stage:}
    The synapse stage preprocesses input events on arrival. The preprocessing results in pre-synaptic weight sums for each neuron, which are stored in dedicated BRAMs. The synapse stage also tracks which neurons have received at least one event in the current timestep, storing this information in the Hot Neurons FIFO.
  \item \textbf{Neuron Stage:}
    The neuron stage updates the states of all hot neurons and queues spiking neurons for spike emission in the axon stage. It currently implements a Forward-Euler \ac{LIF} neuron model, through which state updates caused by leak can be deferred to the arrival time of the next event.
  \item \textbf{Axon Stage:}
  In the axon stage, each spiking neuron triggers the loading and emission of its associated list of post-synaptic events. The amount of post-synaptic connections is programmed per-neuron in a mapping table.
  \item \textbf{Output Stage (TX):}
  For each emitted event, the output stage checks whether the event is destined for a synapse within the core or for an external destination, branching it to the internal feedback path or the external interconnect, respectively. 
\end{itemize}

\noindent The following sections detail key aspects of YANA's implementation:
\\
\textbf{Timestep Progression Scheme:}
The core operates in discrete timesteps and derives the progression of timesteps from its input event workload. Each stage in the processing pipeline sets a load-dependent idle signal that is read by the core's local control logic. Once the whole pipeline is idle, the control logic sets the signal $core_{done}$, which signifies that the core is ready to progress to the next timestep.
Ultimately, an external layer of control logic is responsible for progressing the core's input signal $timestep$, depicted in Figure~\ref{YANA-core-block-diagram}.
In this work's evaluation, we implement timestep progression logic for the use case of processing neuromorphic datasets. Dedicated control logic reads timestamps from the dataset sample's events and progresses the timestep 
accordingly, without any downtime related to elapsed wall time; depending on the workload, this enables faster-than-realtime processing.
\\
\textbf{Input Event Preprocessing Scheme:}
The discretization of timesteps in \acp{SNN} requires that each event is assigned to a specific timestep for processing. YANA's synapse stage preprocesses arriving
input events to completely avoid raw input event buffering while maintaining a steady input rate of $1\nicefrac{\text{event}}{\text{cycle}}$. As events arrive, the synapse stage continuously integrates the associated weights into per-neuron weight sums. A weight sum represents the total (unleaked) input current that will be added to the respective neuron's membrane potential in
the next timestep.
\\
\textbf{Event-driven Neuron Updates:}
YANA updates neuron states only once they are \textit{hot}, that is, once they have received at least one event in the current timestep. This applies both to integrating input currents and to processing continuous leak dynamics. To this end, the neuron stage implements the forward Euler solution of the \ac{LIF} neuron model:
\begin{equation}
  \label{eq:lif_update}
  \tilde{u}(t+n) = u(t) \cdot \left(1 - \frac{1}{\tau_{\text{mem}}}\right)^n + \frac{1}{\tau_{\text{mem}}} \cdot i(t)
\end{equation}
\begin{equation}
  \label{eq:lif_spike_condition}
  u(t+n) = \begin{cases}
    \tilde{u}(t+n) & \text{if } \tilde{u}(t+n) \leq u_{\text{th}} \\
    0 & \text{if } \tilde{u}(t+n) > u_{\text{th}} \text{ (spike occurs)} \\
    0 & \text{if } n > n_{\text{max}} \text{ (round to 0)} \\
  \end{cases}
\end{equation}
In the equations, $u(t)$ is the neuron's membrane potential at timestep $t$, $I(t)$ is the neuron's input current at timestep $t$, and $\tau_{\text{mem}}$ is the membrane time constant. The parameter $n$ represents the number of timesteps that have passed since the last update of the neuron's state. $\tilde{u}(t+n)$ is a temporary variable that holds the result of the leak and input current integration. In YANA, the equations are fully implemented using fixed-point arithmetic.

Note that only the leak term $(1 - \nicefrac{1}{\tau_{\text{mem}}})^n$ depends on the time that has passed since the last neuron update. To efficiently implement this calculation, the neuron stage uses a \ac{LUT} with precalculated solutions to this term for $n_{\text{max}}$ entries, thereby eliminating the need to calculate the power function in hardware. If more than $n_{\text{max}}$ timesteps have passed since the last update, the neuron's membrane potential is rounded to 0 as shown in the second equation.

To support these deferred neuron updates, the neuron stage also maintains a record of the last access-timestamp for each neuron. This enables the core to calculate the number of elapsed timesteps $n$ when a neuron becomes hot, ensuring the correct application of the leak dynamics.
\\
\textbf{Support for Arbitrary SNN Graphs:}
YANA supports arbitrary SNN graphs by implementing point-to-point connections between pairs of neurons in its hardware architecture. The core maintains both pre-synaptic and post-synaptic information in separate memory structures. Pre-synaptic weights are stored in the synapse stage and accessed during event preprocessing, while post-synaptic destinations (dest.) are stored in connection tables accessed by the axon stage. Each outgoing connection is encoded as an event packet with the format:
\begin{equation*}
    \text{[ Dest. Core $(2bit)$ | Dest. Neuron $(10bit)$ | Dest. Synapse $(17bit)$ ]}
\end{equation*}
Since each event packet specifically addresses both target neuron and synapse, a single pre-synaptic weight in the synapse stage can be referenced by multiple connections, enabling native support for weight reuse and arbitrary pruning techniques. 

\subsection{System on Chip FPGA Integration}

\begin{figure}[t]
    \centering
    \includegraphics[width=0.7\textwidth]{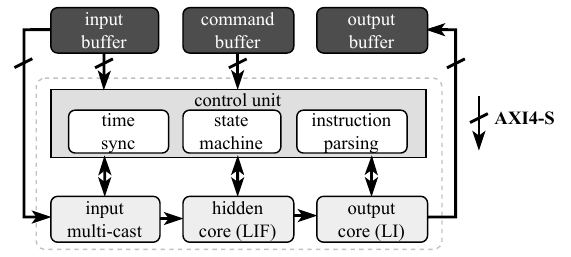}
    \caption{
    YANA architecture integration complete with AXI4 stream buffers, a control unit and cores for input, hidden and output computations.
    }
    \label{fig:exp_setup}
\end{figure}

In this work, we target the low-cost AMD Kria KR260 Robotics Starter Kit. The KR260 employs an AMD Zynq UltraScale+ \ac{SoC} FPGA which integrates software execution in its \ac{PS} and an adaptive FPGA fabric in its \ac{PL}.

As the YANA core architecture by itself only implements the pipelined data flow, there are additional components necessary in order to make it usable in an experimental setup, as can be seen in Figure~\ref{fig:exp_setup}.
To control the execution of the accelerator and communicate with the software stack, a \ac{CU} with a custom command interface is implemented. It keeps track of the global synchronized timestep, decodes and parses incoming commands and maintains an internal state machine to manage the control flow.
To interface with the accelerator's \ac{CU}, we instantiate AXI-enabled buffers for the input data, control commands and result output. These are accessible by the KR260's \ac{PS}, enabling runtime control of the accelerator from software.

Input events are source encoded, only containing information about the source input neuron and the timestep in which the input spike occurred. To multi-cast these input events to the destination encoded format expected by the hidden \ac{LIF} core, an input multi-cast core is integrated. Analogously, an output core containing \ac{LI} neurons integrates all spikes coming from the hidden core. Upon request, it returns the integrated neuron potential back to the \ac{PS} through the output buffer.

We use the previously mentioned dataset sample execution mode for the experiments conducted in this work, which involves loading entire samples into the input buffers of the accelerator and executing them as fast as possible until all input is processed.
After the execution of one sample finishes, the data path including any stateful memories is reset in preparation for the next sample.

\subsection{FPGA Resource Utilization}
Table~\ref{tab:resource-utilization} summarizes the resource utilization of a single YANA core and of the aforementioned experimental deployment on the KR260's \ac{PL}.
Our experimental deployment of YANA on the KR260 utilizes URAMs for storage of synaptic weights and event packets.
Synaptic parameters utilize the largest share of parameter storage in most applications; thus, URAMs are preferred over BRAMs to store them, if available.
Beyond FPGA resource utilization metrics, the table also specifies the resulting storage limits for mappable SNN parameters. YANA's \ac{RTL} design is thoroughly parameterized, and the \ac{SNN}-parameter storage limits are a result of configuring the desired number of synapses and neurons per core pre-synthesis, according to which synthesis infers the required amount of BRAM and URAM resources. Note that our experimental deployment of YANA in this work does not target application specific \ac{PL}-memory resource optimization. 
\begin{table*}
    \centering
    \resizebox{1\textwidth}{!}{
        \begin{threeparttable}
            \caption{FPGA resource utilization and available parameter storage of YANA core and experimental setup on KR260.}
            \label{tab:resource-utilization}
                \begin{tabular}{l*{5}{c}*{4}{c}} 
                    \toprule
                        \multirow{2}{*}[-2pt]{\textbf{Descriptor}} &
                        \multicolumn{5}{c}{\textbf{FPGA Resources}} &
                        \multicolumn{4}{c}{\textbf{Mappable Parameters}} \\
                    \cmidrule(lr){2-6} \cmidrule(lr){7-10}
                        & LUT & Register & BRAM\tnote{a} & URAM & DSP &
                        $syn_{input}$ & $syn_{hidden}$ & $neur_{hidden}$ & $neur_{output}$\\
                    \midrule
                        YANA Core & 740 & 918 & 7 (199) & 24 & 2 & --- & $2^{17}$ & $2^{10}$ & --- \\
                        \text{Deployment\tnote{b} } & 1687 & 1817 & 13.5 (397.5) & 48 & 4 & $2^{17}$ & $2^{17}$ & $2^{10}$ & $2^{10}$ \\
                    \bottomrule
                \end{tabular}
            \begin{tablenotes}
                \footnotesize
                \item[a] Values in brackets represent estimate when URAMs are unavailable. One URAM corresponds approximately to 8 BRAMs.
                \item[b] Excludes AXI IP periphery for integration with PS.
            \end{tablenotes}
        \end{threeparttable}
    }
\end{table*}

\section{YANA Software Framework}
    When developing any hardware architecture, the accompanying software is the key to making
it widely usable without large overheads stemming from custom deployment solutions. Especially
in the \ac{DL} community, there are established tools and workflows that many practitioners
commonly use to develop their neural networks. Therefore, our goal for the software stack
supporting YANA is to provide an end-to-end framework that seamlessly integrates with a wide
range of popular open-source tools for neuromorphic computing.
\begin{figure}[t]
    \centering
    \includegraphics[width=0.9\linewidth]{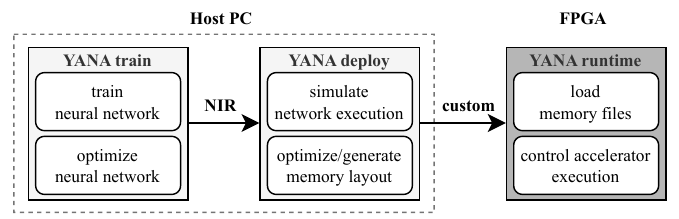}
    \caption{Overview of the YANA software framework}
    \label{fig:yana-sw-framework}
\end{figure}
Notably, we chose the Neuromorphic Intermediate Representation (NIR) \cite{pedersen_neuromorphic_2024}
as our intermediary format throughout the framework to ensure compatibility with other frameworks for
neuromorphic computing.
The YANA software framework aims
to provide an easy-to-use workflow for hardware-aware training, optimization and deployment \acp{SNN}
targeting our custom neuromorphic accelerator.
It consists of three major modules: YANA train, YANA deploy and YANA runtime, as can be seen in Figure~\ref{fig:yana-sw-framework}. The modules are briefly described as follows:
\\
\textbf{YANA train:}
This module facilitates the training of \acp{SNN} for YANA. Currently, it is based
on the Norse framework~\cite{norse2021}, utilizing additional tools like Tonic~\cite{lenz_gregor_2021_5079802} and
PyTorch Lightning~\cite{Falcon_PyTorch_Lightning_2019}.
Its main addition to Norse is an implementation of YANA's quantized fixed-point \ac{LIF} neuron that includes the \ac{LUT}-based leak calculation.
This allows for hardware-aware training, as the quantization effects and \ac{LUT}-based leak can be directly incorporated into the training process.
Note that the current form of YANA train should be considered
a reference implementation for supporting the \ac{DL} programming model on YANA and can be easily exchanged for any other training framework that supports exporting to NIR.
\\
\textbf{YANA deploy:}
This module handles the translation of trained SNN models into hardware-deployable configurations through several key steps. First, it parses the NIR graph representation and simulates it using the accelerator's fixed-point arithmetic to ensure numeric accuracy. The logical network structure is then transformed into an optimized memory layout that maps individual neurons and weights onto the physical hardware resources.
The module outputs memory configuration files for synapse and routing tables in a custom human-readable format that can be directly loaded onto the accelerator. While currently focusing on dense layers, YANA's native suport for weight sharing provides the foundation for future support of convolutional operations and other structured connectivity patterns.
\\
\textbf{YANA runtime:}
Unlike the previous modules which run on a host computer, the runtime executes directly on the \ac{SoC} FPGA's \ac{PS}. It leverages the PYNQ library~\cite{advanced_micro_devices_inc_pynq_2024} to interface with AMD AXI IPs implemented in the \ac{PL} fabric. The runtime initializes the accelerator with the memory configuration files created by YANA deploy, loads input spike events, and orchestrates the execution sequence. During inference, it tracks performance metrics like execution latency, timestep count, and memory utilization.
The current implementation supports sample-based dataset processing, where complete input samples are preloaded into the accelerator's buffer and processed in batch mode. Future development will introduce stream processing capabilities to enable integration with (event-based) sensors for deployment in real-world scenarios and low-power applications.

\section{Evaluation}
    \label{sec:evaluation}
    \subsection{Experiments on Inference Latency Scaling}
\label{subsec:experiments}
In order to test the latency scaling of YANA's fully event-driven \acp{SNN} acceleration, by using our software framework, we prepare networks and data samples of different spatial and temporal sparsity levels and measure their inference time after deployment on the FPGA. 
For this, we train networks with one hidden layer of 100 \ac{LIF} units on the Spiking Heidelberg Digits (SHD) dataset.
Subsequently, we gradually increase temporal sparsity levels of the input data by uniformly dropping events. Likewise, we increase the spatial sparsity of the feed-forward layers by pruning certain percentages of weights ranking lowest in magnitude.
Note that, while we initially trained the networks until convergence to obtain realistic activity patterns, we do not take the accuracy of the networks into account. YANA's fixed-point \ac{LIF} implementation is numerically stable. Any task accuracy achieved during training in simulation is therefore preserved during inference on YANA.
In fact, the networks are meant to show the ability of YANA's architecture to leverage spatial and temporal sparsity for efficiency gains in network execution, rather than serve as a benchmark for optimally solving the SHD task.
\subsection{Results}
\begin{figure*}[t]
    \centering
    \includegraphics[width=\linewidth]{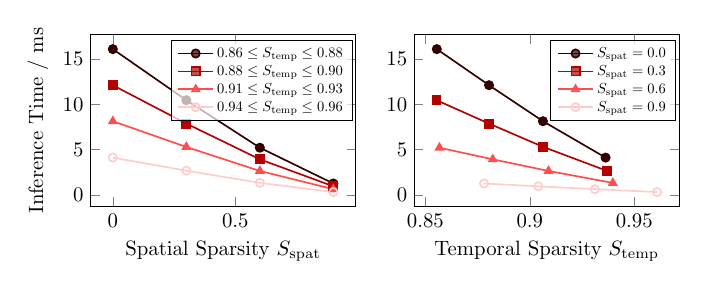}
    \caption{
    Scaling of inference time of \acp{SNN} with different spatial and temporal sparsity levels after deployment on YANA. 
    In the case of sweeping $S_\text{spat}$, $S_\text{temp}$ cannot be fixed and has to be given in a small range, since the pruning level influences hidden layer sparsity, ultimately changing the total $S_\text{temp}$.
    }
    \label{fig:exp_results}
\end{figure*}
Figure \ref{fig:exp_results} displays the results of the experiments described above. The inference time of the deployed \acp{SNN} is computed as the average of 20 samples executed on the accelerator hardware clocked at 100 MHz.
It decreases almost linearly with both increasing spatial and temporal sparsity levels of weights and data.
Since all incoming events are multicast to all connected weights, the total number of events to be processed is given by the sum of non-zero weight counts downstream of each event.
Therefore, reducing either the number of weights or the number of input events directly impacts the total number of events to be processed, effectively reducing the latency caused by the synapse and axon subcores.
Additionally, for smaller network sizes and large levels of sparsity, we observe that the inference time plateaus.
We explain this through ever fewer computation being performed in the synapse and axon cores, emphasizing the static latency ceiling of the neuron sub module and of the core control logic.
Notably, while recordings in the SHD dataset are predominantly 500ms to 1000ms long, inference times in our experiments range from 16ms to 1ms, demonstrating the speed of YANA's event-driven processing pipeline. The results showcase the considerable impact of sparsity on inference time, underpinning the importance of spike regularization and pruning techniques in \ac{SNN} application development.

\section{Discussion \& Future Work}
    While YANA demonstrates significant potential for bridging the neuromorphic simulation-to-hardware gap, several limitations and opportunities for improvement warrant discussion.

\newpage\noindent\textbf{Limited Layer Type Support:}
At the time of this work's experiments, \textit{YANA deploy} could only map feed forward layers onto YANA. However, many practical \ac{SNN} applications benefit from layers that exploit weight reuse patterns, such as convolutional layers. YANA's architecture fundamentally supports weight reuse and future development will extend the software framework to support convolutional layers and other structured connectivity patterns.
\\
\textbf{Event-Driven Power Efficiency Validation:}
While event-driven processing is a fundamental design principle of YANA, the actual power savings achieved through this approach on FPGA hardware have yet to be empirically demonstrated. Power efficiency represents a major argument for the Neuromorphic Advantage, and quantifying it is crucial for validating YANA's effectiveness. Future work will conduct detailed power characterization studies using load-dependent power saving strategies such as clock gating and SRAM sleep modes.
\\
\textbf{Scalability and Many-Core Deployment:}
The current evaluation focuses on an experimental deployment of YANA with an input (multicast), hidden and output core. However, YANA is designed with scalability in mind, and future research will investigate parallel processing speedups and input event rate improvements when YANA is deployed as a many-core architecture. This will involve deploying YANA in a \ac{NoC} setup to enable inter-core communication and exploring load balancing strategies for distributing \ac{SNN} processing across multiple cores.

\section{Conclusion}
    This work introduces YANA, an open-source neuromorphic accelerator designed to bridge the simulation-to-hardware gap that currently limits the advancement of \ac{SNN} research and development. By providing accessible neuromorphic computing capabilities on affordable FPGA platforms, YANA enables researchers and practitioners to move beyond simulation-only SNN development and validate their algorithms on actual event-driven hardware.

YANA's architecture demonstrates several key contributions to the neuromorphic computing field. Its event-driven processing pipeline efficiently exploits both temporal and spatial sparsity inherent in SNNs, achieving near-linear scaling of inference time with sparsity levels.
Furthermore, YANA's support for arbitrary SNN topologies through point-to-point connections is future proof for the evolving landscape of neuromorphic applications, allowing for flexible network designs that can adapt to various research needs.

Our experimental evaluation on the Spiking Heidelberg Digits dataset validates YANA's ability to leverage sparsity for performance improvements, while the deployment on the accessible AMD Kria KR260 platform demonstrates practical resource efficiency with manageable hardware requirements. The YANA software framework, from training through deployment, provides a complete solution that integrates with the existing neuromorphic open-source ecosystem.
As the neuromorphic computing field continues to mature, accessible platforms like YANA will be crucial for translating the promising theoretical advances in SNNs into practical, real-world applications that fully exploit the unique advantages of brain-inspired computing paradigms.

\bibliographystyle{splncs04}
\bibliography{references.bib}

@article{diamond_comparing_2016,
	title = {Comparing {Neuromorphic} {Solutions} in {Action}: {Implementing} a {Bio}-{Inspired} {Solution} to a {Benchmark} {Classification} {Task} on {Three} {Parallel}-{Computing} {Platforms}},
	volume = {9},
	issn = {1662-453X},
	shorttitle = {Comparing {Neuromorphic} {Solutions} in {Action}},
	doi = {10.3389/fnins.2015.00491},
	urldate = {2025-05-27},
	journal = {Frontiers in Neuroscience},
	author = {Diamond, Alan and Nowotny, Thomas and Schmuker, Michael},
	month = jan,
	year = {2016},
}

@inproceedings{aimone_provable_2021,
	address = {Virtual Event USA},
	title = {Provable {Advantages} for {Graph} {Algorithms} in {Spiking} {Neural} {Networks}},
	isbn = {978-1-4503-8070-6},
	doi = {10.1145/3409964.3461813},
	language = {en},
	urldate = {2025-05-28},
	booktitle = {Proceedings of the 33rd {ACM} {Symposium} on {Parallelism} in {Algorithms} and {Architectures}},
	publisher = {ACM},
	author = {Aimone, James B. and Ho, Yang and Parekh, Ojas and Phillips, Cynthia A. and Pinar, Ali and Severa, William and Wang, Yipu},
	month = jul,
	year = {2021},
	pages = {35--47},
}

@article{frenkel_bottom-up_2023,
	title = {Bottom-{Up} and {Top}-{Down} {Approaches} for the {Design} of {Neuromorphic} {Processing} {Systems}: {Tradeoffs} and {Synergies} {Between} {Natural} and {Artificial} {Intelligence}},
	volume = {111},
	copyright = {https://ieeexplore.ieee.org/Xplorehelp/downloads/license-information/IEEE.html},
	issn = {0018-9219, 1558-2256},
	shorttitle = {Bottom-{Up} and {Top}-{Down} {Approaches} for the {Design} of {Neuromorphic} {Processing} {Systems}},
	doi = {10.1109/JPROC.2023.3273520},
	number = {6},
	urldate = {2025-05-26},
	journal = {Proceedings of the IEEE},
	author = {Frenkel, Charlotte and Bol, David and Indiveri, Giacomo},
	month = jun,
	year = {2023},
	pages = {623--652},
}

@article{kudithipudi_neuromorphic_2025,
	title = {Neuromorphic computing at scale},
	volume = {637},
	issn = {0028-0836, 1476-4687},
	doi = {10.1038/s41586-024-08253-8},
	language = {en},
	number = {8047},
	urldate = {2025-05-27},
	journal = {Nature},
	author = {Kudithipudi, Dhireesha and Schuman, Catherine and Vineyard, Craig M. and Pandit, Tej and Merkel, Cory and Kubendran, Rajkumar and Aimone, James B. and Orchard, Garrick and Mayr, Christian and Benosman, Ryad and Hays, Joe and Young, Cliff and Bartolozzi, Chiara and Majumdar, Amitava and Cardwell, Suma George and Payvand, Melika and Buckley, Sonia and Kulkarni, Shruti and Gonzalez, Hector A. and Cauwenberghs, Gert and Thakur, Chetan Singh and Subramoney, Anand and Furber, Steve},
	month = jan,
	year = {2025},
	pages = {801--812},
}

@article{muir_road_2025,
	title = {The road to commercial success for neuromorphic technologies},
	volume = {16},
	issn = {2041-1723},
	doi = {10.1038/s41467-025-57352-1},
	language = {en},
	number = {1},
	urldate = {2025-05-27},
	journal = {Nature Communications},
	author = {Muir, Dylan Richard and Sheik, Sadique},
	month = apr,
	year = {2025},
	pages = {3586},
}

@article{yik_neurobench_2025,
	title = {The neurobench framework for benchmarking neuromorphic computing algorithms and systems},
	volume = {16},
	issn = {2041-1723},
	doi = {10.1038/s41467-025-56739-4},
	language = {en},
	number = {1},
	urldate = {2025-05-27},
	journal = {Nature Communications},
	author = {Yik, Jason and Van Den Berghe, Korneel and Den Blanken, Douwe and Bouhadjar, Younes and Fabre, Maxime and Hueber, Paul and Ke, Weijie and Khoei, Mina A. and Kleyko, Denis and Pacik-Nelson, Noah and Pierro, Alessandro and Stratmann, Philipp and Sun, Pao-Sheng Vincent and Tang, Guangzhi and Wang, Shenqi and Zhou, Biyan and Ahmed, Soikat Hasan and Vathakkattil Joseph, George and Leto, Benedetto and Micheli, Aurora and Mishra, Anurag Kumar and Lenz, Gregor and Sun, Tao and Ahmed, Zergham and Akl, Mahmoud and Anderson, Brian and Andreou, Andreas G. and Bartolozzi, Chiara and Basu, Arindam and Bogdan, Petrut and Bohte, Sander and Buckley, Sonia and Cauwenberghs, Gert and Chicca, Elisabetta and Corradi, Federico and De Croon, Guido and Danielescu, Andreea and Daram, Anurag and Davies, Mike and Demirag, Yigit and Eshraghian, Jason and Fischer, Tobias and Forest, Jeremy and Fra, Vittorio and Furber, Steve and Furlong, P. Michael and Gilpin, William and Gilra, Aditya and Gonzalez, Hector A. and Indiveri, Giacomo and Joshi, Siddharth and Karia, Vedant and Khacef, Lyes and Knight, James C. and Kriener, Laura and Kubendran, Rajkumar and Kudithipudi, Dhireesha and Liu, Shih-Chii and Liu, Yao-Hong and Ma, Haoyuan and Manohar, Rajit and Margarit-Taulé, Josep Maria and Mayr, Christian and Michmizos, Konstantinos and Muir, Dylan R. and Neftci, Emre and Nowotny, Thomas and Ottati, Fabrizio and Ozcelikkale, Ayca and Panda, Priyadarshini and Park, Jongkil and Payvand, Melika and Pehle, Christian and Petrovici, Mihai A. and Posch, Christoph and Renner, Alpha and Sandamirskaya, Yulia and Schaefer, Clemens J. S. and Van Schaik, André and Schemmel, Johannes and Schmidgall, Samuel and Schuman, Catherine and Seo, Jae-sun and Sheik, Sadique and Shrestha, Sumit Bam and Sifalakis, Manolis and Sironi, Amos and Stewart, Kenneth and Stewart, Matthew and Stewart, Terrence C. and Timcheck, Jonathan and Tömen, Nergis and Urgese, Gianvito and Verhelst, Marian and Vineyard, Craig M. and Vogginger, Bernhard and Yousefzadeh, Amirreza and Zohora, Fatima Tuz and Frenkel, Charlotte and Reddi, Vijay Janapa},
	month = feb,
	year = {2025},
	pages = {1545},
}

@misc{bos_sub-mw_2022,
	title = {Sub-{mW} {Neuromorphic} {SNN} audio processing applications with {Rockpool} and {Xylo}},
	doi = {10.48550/arXiv.2208.12991},
	urldate = {2025-06-12},
	publisher = {arXiv},
	author = {Bos, Hannah and Muir, Dylan},
	month = sep,
	year = {2022},
	note = {arXiv:2208.12991 [cs]},
}

@inproceedings{shrestha_efficient_2024,
	address = {Seoul, Korea, Republic of},
	title = {Efficient {Video} and {Audio} {Processing} with {Loihi} 2},
	copyright = {https://doi.org/10.15223/policy-029},
	isbn = {9798350344851},
	doi = {10.1109/ICASSP48485.2024.10448003},
	urldate = {2025-06-12},
	booktitle = {{ICASSP} 2024 - 2024 {IEEE} {International} {Conference} on {Acoustics}, {Speech} and {Signal} {Processing} ({ICASSP})},
	publisher = {IEEE},
	author = {Shrestha, Sumit Bam and Timcheck, Jonathan and Frady, Paxon and Campos-Macias, Leobardo and Davies, Mike},
	month = apr,
	year = {2024},
	pages = {13481--13485},
}

@misc{hoppner_spinnaker_2022,
	title = {The {SpiNNaker} 2 {Processing} {Element} {Architecture} for {Hybrid} {Digital} {Neuromorphic} {Computing}},
	doi = {10.48550/arXiv.2103.08392},
	urldate = {2025-06-12},
	publisher = {arXiv},
	author = {Höppner, Sebastian and Yan, Yexin and Dixius, Andreas and Scholze, Stefan and Partzsch, Johannes and Stolba, Marco and Kelber, Florian and Vogginger, Bernhard and Neumärker, Felix and Ellguth, Georg and Hartmann, Stephan and Schiefer, Stefan and Hocker, Thomas and Walter, Dennis and Liu, Genting and Garside, Jim and Furber, Steve and Mayr, Christian},
	month = aug,
	year = {2022},
	note = {arXiv:2103.08392 [cs]},
}

@article{richter_dynap-se2_2024,
	title = {{DYNAP}-{SE2}: a scalable multi-core dynamic neuromorphic asynchronous spiking neural network processor},
	volume = {4},
	issn = {2634-4386},
	shorttitle = {{DYNAP}-{SE2}},
	doi = {10.1088/2634-4386/ad1cd7},
	number = {1},
	urldate = {2025-06-12},
	journal = {Neuromorphic Computing and Engineering},
	author = {Richter, Ole and Wu, Chenxi and Whatley, Adrian M and Köstinger, German and Nielsen, Carsten and Qiao, Ning and Indiveri, Giacomo},
	month = mar,
	year = {2024},
	pages = {014003},
}

@article{carpegna_spiker_2024,
	title = {Spiker+: a framework for the generation of efficient {Spiking} {Neural} {Networks} {FPGA} accelerators for inference at the edge},
	copyright = {https://creativecommons.org/licenses/by/4.0/legalcode},
	issn = {2168-6750, 2376-4562},
	shorttitle = {Spiker+},
	doi = {10.1109/TETC.2024.3511676},
	urldate = {2025-06-13},
	journal = {IEEE Transactions on Emerging Topics in Computing},
	author = {Carpegna, Alessio and Savino, Alessandro and Carlo, Stefano Di},
	year = {2024},
	pages = {1--15},
}

@article{leone_syntzulu_2025,
	title = {{SYNtzulu}: {A} {Tiny} {RISC}-{V}-{Controlled} {SNN} {Processor} for {Real}-{Time} {Sensor} {Data} {Analysis} on {Low}-{Power} {FPGAs}},
	volume = {72},
	copyright = {https://creativecommons.org/licenses/by/4.0/legalcode},
	issn = {1549-8328, 1558-0806},
	shorttitle = {{SYNtzulu}},
	doi = {10.1109/TCSI.2024.3450966},
	number = {2},
	urldate = {2025-06-13},
	journal = {IEEE Transactions on Circuits and Systems I: Regular Papers},
	author = {Leone, Gianluca and Antonio Scrugli, Matteo and Badas, Lorenzo and Martis, Luca and Raffo, Luigi and Meloni, Paolo},
	month = feb,
	year = {2025},
	pages = {790--801},
}

@article{saulquin_modnef_2025,
	title = {{ModNEF} : {An} {Open} {Source} {Modular} {Neuromorphic} {Emulator} for {FPGA} for {Low}-{Power} {In}-{Edge} {Artificial} {Intelligence}},
	issn = {1544-3566, 1544-3973},
	shorttitle = {{ModNEF}},
	doi = {10.1145/3730581},
	language = {en},
	urldate = {2025-06-13},
	journal = {ACM Transactions on Architecture and Code Optimization},
	author = {Saulquin, Aurélie and Fatahi, Mazdak and Boulet, Pierre and Meftali, Samy},
	month = apr,
	year = {2025},
	pages = {3730581},
}

@misc{davies_taking_2021,
	title = {Taking {Neuromorphic} {Computing} to the {Next} {Level} with {Loihi} 2},
	language = {en},
	author = {Davies, Mike},
	month = sep,
	year = {2021},
}

@misc{Falcon_PyTorch_Lightning_2019,
	title = {{PyTorch} lightning},
	copyright = {Apache-2.0},
	url = {https://github.com/Lightning-AI/lightning},
	author = {Falcon, William and {The PyTorch Lightning team}},
	month = mar,
	year = {2019},
	doi = {10.5281/zenodo.3828935},
}

@misc{lenz_gregor_2021_5079802,
	title = {Tonic: event-based datasets and transformations.},
	url = {https://doi.org/10.5281/zenodo.5079802},
	publisher = {Zenodo},
	author = {Lenz, Gregor and Chaney, Kenneth and Shrestha, Sumit Bam and Oubari, Omar and Picaud, Serge and Zarrella, Guido},
	month = jul,
	year = {2021},
	doi = {10.5281/zenodo.5079802},
}

@misc{advanced_micro_devices_inc_pynq_2024,
	title = {{PYNQ} - {Python} {Productivity} for {AMD} {Adaptive} {Computing} {Platforms}},
	url = {https://www.pynq.io/},
	author = {{Advanced Micro Devices, Inc.}},
	month = feb,
	year = {2024},
}

@misc{norse2021,
	title = {Norse - {A} deep learning library for spiking neural networks},
	url = {https://doi.org/10.5281/zenodo.4422025},
	publisher = {Zenodo},
	author = {Pehle, Christian and Pedersen, Jens Egholm},
	month = jan,
	year = {2021},
	doi = {10.5281/zenodo.4422025},
}

@article{pedersen_neuromorphic_2024,
	title = {Neuromorphic intermediate representation: {A} unified instruction set for interoperable brain-inspired computing},
	volume = {15},
	copyright = {2024 The Author(s)},
	issn = {2041-1723},
	shorttitle = {Neuromorphic intermediate representation},
	url = {https://www.nature.com/articles/s41467-024-52259-9},
	doi = {10.1038/s41467-024-52259-9},
	language = {en},
	number = {1},
	urldate = {2025-05-28},
	journal = {Nature Communications},
	author = {Pedersen, Jens E. and Abreu, Steven and Jobst, Matthias and Lenz, Gregor and Fra, Vittorio and Bauer, Felix Christian and Muir, Dylan Richard and Zhou, Peng and Vogginger, Bernhard and Heckel, Kade and Urgese, Gianvito and Shankar, Sadasivan and Stewart, Terrence C. and Sheik, Sadique and Eshraghian, Jason K.},
	month = sep,
	year = {2024},
	note = {Publisher: Nature Publishing Group},
	pages = {8122},
}

@article{cramer_heidelberg_2022,
	title = {The {Heidelberg} {Spiking} {Data} {Sets} for the {Systematic} {Evaluation} of {Spiking} {Neural} {Networks}},
	volume = {33},
	copyright = {https://creativecommons.org/licenses/by/4.0/legalcode},
	issn = {2162-237X, 2162-2388},
	url = {https://ieeexplore.ieee.org/document/9311226/},
	doi = {10.1109/TNNLS.2020.3044364},
	language = {en},
	number = {7},
	urldate = {2025-05-27},
	journal = {IEEE Transactions on Neural Networks and Learning Systems},
	author = {Cramer, Benjamin and Stradmann, Yannik and Schemmel, Johannes and Zenke, Friedemann},
	month = jul,
	year = {2022},
	pages = {2744--2757},
}
\end{document}